\documentclass[conference]{IEEEtran}
\IEEEoverridecommandlockouts
\usepackage{cite}
\usepackage{amsmath,amssymb,amsfonts}
\usepackage{algorithmic}
\usepackage{graphicx}
\usepackage{textcomp}
\usepackage{xcolor}
\usepackage{enumitem}
\def\BibTeX{{\rm B\kern-.05em{\sc i\kern-.025em b}\kern-.08em
    T\kern-.1667em\lower.7ex\hbox{E}\kern-.125emX}}
\begin{document}

\title{Embedding Ordinality to Binary Loss Function for Improving Solar Flare Forecasting}
\author{\IEEEauthorblockN{Chetraj Pandey, Anli Ji, Jinsu Hong, Rafal A. Angryk, Berkay Aydin}
\IEEEauthorblockA{\textit{Dept. of Computer Science, Georgia State University, Atlanta, GA, USA} \\
\textit {\{cpandey1, aji1, jhong36, rangryk, baydin2\}@gsu.edu}}
}
\maketitle

\begin{abstract}

Several natural phenomena, such as floods, earthquakes, volcanic eruptions, or extreme space weather events often come with severity indexes. While these indexes, whether linear or logarithmic are vital, data-driven predictive models for these events rather use a fixed threshold. In this paper, we explore encoding this ordinality to enhance the performance of data-driven models, with specific application in solar flare forecasting. The prediction of solar flares is commonly approached as a binary forecasting problem, categorizing events as either Flare (FL) or No-Flare (NF) based on a chosen threshold (e.g., $\geq$C-class, $\geq$M-class, or $\geq$X-class). However, this binary formulation overlooks the inherent ordinality between the sub-classes within each binary class (FL and NF). In this paper, we propose a novel loss function aimed at optimizing the binary flare prediction problem by embedding the intrinsic ordinal flare characteristics into the binary cross-entropy (BCE) loss function. This modification is intended to provide the model with better guidance based on the ordinal characteristics of the data and improve the overall performance of the models. For our experiments, we employ a ResNet34-based model with transfer learning to predict $\geq$M-class flares by utilizing the shape-based features of magnetograms of active region (AR) patches spanning from $-$90$^{\circ}$ to $+$90$^{\circ}$ of solar longitude as our input data. We use a composite skill score (CSS) as our evaluation metric, which is calculated as the geometric mean of the True Skill Score (TSS) and the Heidke Skill Score (HSS) to rank and compare our models' performance. The primary contributions of this work are as follows: (i) We introduce a novel approach to encode ordinality into a binary loss function showing an application to solar flare prediction, (ii) We enhance solar flare forecasting by enabling flare predictions for each AR across the entire solar disk, without any longitudinal restrictions, and evaluate and compare performance. (iii) Our candidate model, optimized with the proposed loss function, shows an improvement of $\sim$7\%, $\sim$4\%, and $\sim$3\% for AR patches within $\pm$30$^\circ$, $\pm$60$^\circ$, and $\pm$90$^\circ$ of solar longitude, respectively in terms of CSS, when compared with standard BCE. Additionally, we demonstrate the ability to issue flare forecasts for ARs in near-limb regions (regions between $\pm$60$^{\circ}$ to $\pm$90$^{\circ}$) with a CSS=0.34 (TSS=0.50 and HSS=0.23), expanding the scope of AR-based models for solar flare prediction. This advances the reliability of solar flare forecasts, leading to more effective prediction capabilities.
\end{abstract}


\section{Introduction}
From earthquakes to tornadoes, and volcanic eruptions to extreme space weather events, natural occurrences that pose hazards to our society often come with a severity index. This index may follow a linear scale (such as flood severity \cite{flood} or tornadoes \cite{tornadoes}) or a logarithmic one (for example, earthquakes \cite{earthquake}, volcanic activity \cite{volcanoes}, or space weather events like flares or solar energetic particle events \cite{spaceweather}). Predictive models for these events commonly incorporate a set threshold; however, these models can gain advantages from incorporating the ordinality of these severity indices. In this work, we will delve into encoding this information appropriately to efficiently optimize data-driven predictive models, with specific applications to binary solar flare forecasting.  

Solar flares are short-lived events on the Sun observed as intense outbursts of energy radiating from the Sun's surface in the form of extreme ultraviolet and X-ray radiation, and they are the central phenomena in space weather forecasting. They are classified according to their peak X-ray flux level into the following five categories by the National Oceanic and Atmospheric Administration (NOAA):  X $(>10^{-4}Wm^{-2})$, M $(>10^{-5}Wm^{-2})$, C $(>10^{-6}Wm^{-2})$, B $(>10^{-7}Wm^{-2})$, and A $(>10^{-8}Wm^{-2})$ \cite{spaceweather}. These five major flare classes are measured on a logarithmic scale and ordered as X$>$M$>$C$>$B$>$A. Flares weaker than A-class are generally undetectable and are classified as flare-quiet (FQ). M-class and X-class solar flares are rare events and much more powerful than other flare classes. These stronger flares (M- and X-class) attract the attention of researchers because they can potentially impact conditions near Earth and disrupt technological systems such as satellite communications, GPS navigation, power grids, and aviation \cite{Yasyukevich2018}. Therefore, solar flare prediction in a binary setting is most commonly formulated as predicting $\geq$M-class flares. \\

\noindent\textbf{Ordinality-aware Loss Function: } 
In solar flare forecasting, the binary prediction framework involves categorizing flares based on their flare magnitude. Specifically, setting the threshold at $\geq$M categorizes M- and X-class flares as Flare (FL), while FQ-, A-, B-, and C-class flares are designated as No Strong Flare (NF). This approach simplifies prediction by distinguishing significant flares (M- and X-class) from less intense activity, aiding in assessing potential solar disturbances. However, the intrinsic ordinal flare characteristics in sub-class level is overlooked during model optimization. Traditional loss functions like cross-entropy and focal loss \cite{focal},  commonly used in data-driven model optimization for binary settings, cannot account for ordinal characteristics within the FL and NF classes. They treat all instances equally during optimization, failing to distinguish between different sub-classes within these categories. This approach does not fully utilize the ordinal information inherent in the flare classification system. Hence, in our work, we propose encoding these ordinal characteristics as weighting factors in the binary cross-entropy (BCE) loss function. By doing so, we assign different weights to instances based on their flare sub-class, ensuring that the model optimizes for the specific nuances within each class. We hypothesize that this adjustment is particularly relevant considering the ordinal nature of the flare events classification.\\

\noindent\textbf{Flare Forecasting with Projection Effects: }
In this study, we utilize images of line-of-sight (LoS) magnetograms of active region (AR) patches.  ARs are high activity regions on the Sun's surface, distinguished by their intense magnetic fields concentrated within sunspots. These magnetic fields often undergo significant distortion and instability, triggering plasma disturbances and releasing energy in the form of flares and other solar phenomena \cite{Toriumi2019}). This makes ARs the regions of interest emphasizing the importance of utilizing AR-based features for predicting solar flares, as the disturbed magnetic fields in them are often linked as the main initiators of these solar events. However, the magnetic field measurements, which are the main features used in AR-based forecasting techniques, are susceptible to severe projection effects caused by the orientation of the observing instrument relative to the solar surface. Therefore, as ARs get closer to limbs to the degree that after $\pm$60$^{\circ}$ of solar longitude, the magnetic field readings are distorted \cite{Falconer2016}, which limits the existing models to include data pertaining to central locations only \cite{pandeyecml2023}. To address this, we derive images from original LoS magnetogram rasters of AR patches.\\

\noindent\textbf{Data Preprocessing Pipeline for ARs: }
Our data preprocessing pipeline introduced in \cite{pandeyecml2024}, converts high-dimensional magnetic field rasters to images capturing the overall morphology and spatial distribution of ARs retaining the important shape-based parameters such as size, directionality, sunspot borders, and polarity inversion lines \cite{ji2023systematic}. Shape-based features retained in these derived images of magnetograms provide a robust representation of the overall underlying magnetic activity. We recognize the persistence of severe projection effects, however, we hypothesize that the complex feature learning capabilities of contemporary deep learning models can potentially learn from these distorted readings. Consequently, we include data encompassing ARs in near-limb regions (beyond $\pm$60$^{\circ}$) as well, thereby offering a novel capability to predict solar flares throughout the entire solar disk. 

Furthermore, it is essential to note that the tracked AR patches vary in size depending on the size of the ARs. Existing approaches have been limited to AR patches in central locations, often resizing rectangular patches to obtain square images. However, this resizing distorts the original aspect ratio, consequently altering the shapes and sizes of ARs. Alternatively, variable-sized AR patches are cropped (using methods like center crop or random crop) to obtain square images, resulting in information loss. In contrast, we proposed and utilized a sliding window kernel-based approach in \cite{pandeyecml2024}. This method select such a cropped region that maximizes total unsigned flux (USFLUX: the sum of the absolute of the magnetic field strength values), maintaining the original aspect ratios of AR patches and preserving critical spatial features. By maximizing the USFLUX, we ensure that we extract the most representative region with significant magnetic flux build up. This method adapts to the variability in AR patch shapes and sizes, avoiding distortion and prioritizing the capture of more relevant information. 

Leveraging the images of LoS magnetograms of AR patches, we develop a predictive model for solar flares of magnitude $\geq$M-class. We employ these images to train  ResNet34 \cite{resnet} based model with different configurations of our proposed loss function. Our contributions can be summarized as follows: (i) We introduce a novel approach to encode inherent ordinality of data to binary cross-entropy (BCE) loss function, showing the effectiveness in solar flare prediction as a case study,  (ii) We show that our models are capable to predict flares across the entire solar disk, including often overlooked near-limb regions, improving the comprehensiveness of AR-based solar flare prediction models. This study presents a key advancement in the field of solar flare prediction, contributing to ongoing efforts aimed at enhancing space weather forecasting capabilities and improving our understanding of solar phenomena.

The remainder of the paper is structured as follows. Sec.~\ref{sec:rel} provides an overview of existing studies on solar flare predictions using deep learning models and various data sources. In Sec.~\ref{sec:data}, we detail the process of data collection with labeling and consequent data distribution, and describe the architecture of our flare prediction model. In Sec.~\ref{sec:method} we outline our methodology by providing a detailed description of our modification to the standard cross-entropy loss and its application to solar flare prediction. Sec.~\ref{sec:expt} presents the experimental design and hyperparameter configurations of our model. Sec.~\ref{sec:result} presents our model evaluation showing the effectiveness of our approach on models' performance evaluated with skill scores. Finally, in Sec.~\ref{sec:conc}, we summarize our findings and suggest avenues for future research.

\section{Related Work}\label{sec:rel}
Several approaches such as human-based predictions (e.g., \cite{Crown2012}), statistical models (e.g., \cite{Lee2012}), and numerical simulations based on physics-based models (e.g., \cite{Kusano2020}), have been employed to predict solar flares. Recently, the use of data-driven approaches, which leverage machine learning and deep learning techniques, has significantly increased (e.g., \cite{Hong2023, PandeyICMLA2023, Hong2023CogMi, HongICMLA2023}) owing to their capacity to exploit extensive datasets and their experimental achievements in space weather forecasting \cite{Whitman2022}. As solar flares are phenomena caused by sudden, abrupt changes in the magnetic field in the solar atmosphere, these data-driven approaches most commonly utilize magnetogram-based data  which includes solar full-disk magnetograms (e.g., \cite{pandeyecml2023, Pandey2022f, Pandey2023DS, Pandey2023DSAA, PandeyAIKE2023}), multivariate time series (MVTS) data extracted from solar vector magnetograms (e.g., \cite{Ji2020}, \cite{Ji2022}, \cite{Ji2023}), cutouts or patches of tracked AR (e.g., \cite{Huang2018}, \cite{Li2020}), and features summarizing each AR patch (e.g., \cite{Bloomfield2012}, \cite{Bobra2015}).

A deep learning model based on a multi-layer perceptron to predict solar flares $\geq$C and $\geq$M class  was presented in \cite{Nishizuka2018}. In this study, they used manually selected features extracted from multi-modal solar observations of the full solar disk, which included vector magnetograms and extreme ultraviolet (EUV) images to predict $\geq$M- and $\geq$C-class flares.  In \cite{Park2018}, a convolutional neural network (CNN) based hybrid model is proposed to predict the occurrence of $\geq$C-class flares. Similarly, \cite{Pandey2022, Pandey2023DSAA} presented a CNN-based model to predict $\geq$M-class flares utilizing full-disk magnetogram images. While these full-disk models include near-limb regions, by design they are unable to localize the relevant ARs that are likely to flare and instead issue one single forecast for the entire solar disk. 

In \cite{Bobra2015}, a support vector machine based model trained with 25 AR summary parameters extracted from vector magnetograms of AR patches within $\pm$68$^\circ$ of solar longitude was presented. Similarly, in \cite{Ji2022}, a deep learning based time series classifier and in \cite{Ji2023} a sliding window Time Series Forest (TSF)  was trained with a MVTS data of 24 space weather related physical parameters primarily calculated from AR magnetograms within $\pm$70$^\circ$ of solar longitude. Furthermore, a CNN-based flare forecasting model trained with solar AR patches (resized to 100$\times$100 pixels) extracted from LoS magnetograms within $\pm$30$^{\circ}$ of solar longitude to predict $\geq$C-, $\geq$M-, and $\geq$X-class flares was presented in \cite{Huang2018}. More recently, \cite{Li2023} proposed a CNN-based model named ``CARFFM-4'' trained with AR patches (sized to 160$\times$160 pixels) created from R parameter \cite{Schrijver2007} within $\pm$30$^{\circ}$ of solar longitude to predict $\geq$M-class flares in the next 48 hours. 

All the literature reviewed in this section, formulates solar flare prediction as a binary forecasting problem utilizing a binary loss function without any flare ordinal characteristics. Furthermore, It is important to note that, there is variability in the literature in terms of the type of data modality which includes multiple instruments (HMI/SDO, AIA/SDO, MDI/SOHO) and data types (EUV images, magnetograms and extracted features corresponding to AR and full-disk). Furthermore the variability in prediction targets ($\geq$C-, $\geq$M-, $\geq$X-class flares) and forecasting horizon (24 hours and 48 hours) is also prominent. The predictive capabilities of AR-based models are often limited by observations taken from central locations from $\pm30^{\circ}$ to $\pm70^{\circ}$. The full-disk models complement the issue of longitudinal coverage in AR-based models; however, they fail to pin-point an active region and issue a single forecast for the entire solar disk.  In this work, we introduce a new loss function to build a limb-to-limb flare prediction model that is trained on magnetogram images of AR-patches spanning full 180$^{\circ}$ ($\pm90^{\circ}$) of solar longitude and evaluate our models' efficacy in different zones defined by longitudinal range and provide a novel capability, to our knowledge, missing in operational systems.  

\section{Data and Model}\label{sec:data}
\begin{figure}[htbp]
\centering
\includegraphics[width=0.98\linewidth ]{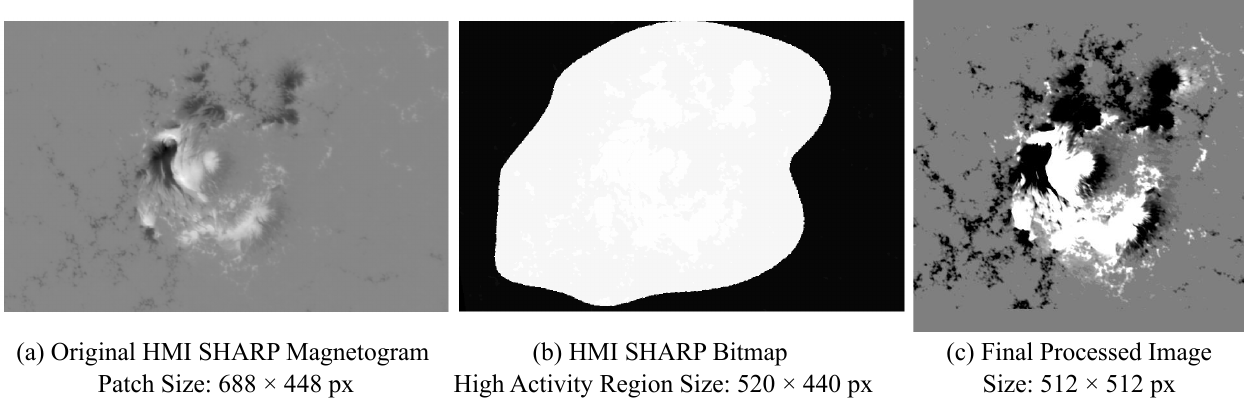}
\caption[]{An illustrative example of (a) Original raw input magnetogram of HMI AR patch corresponding to HARP number: 7115 (NOAA AR number: 12673) observed on 2017-09-06 at 06:00:00 UTC, (b) Bitmap corresponding to HMI AR patch in (a) showing the high activity region (region of interest) indicated by white pixels, (c) Final processed image of AR patch in (a) now sized to 512$\times$512, that is used to train our models.}
\vspace{-10pt}
\label{fig:dataexample}
\end{figure}

\begin{figure*}[htbp]
\centering
\includegraphics[width=0.95\linewidth ]{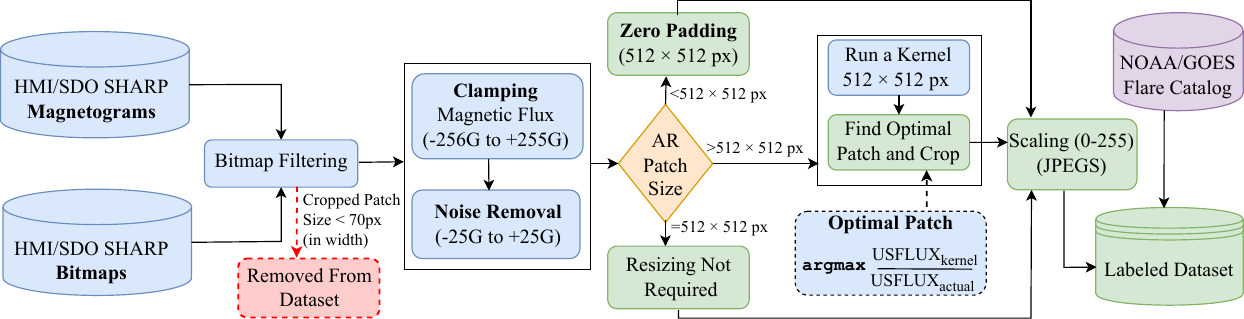}
\caption[]{The overall schema of the data processing pipeline used in this work. It shows a sequential pipeline for creating JPEG images from magnetogram rasters and corresponding bitmaps used for cropping the regions with relevant information. Boxes colored in green collectively defines our dataset.}
\label{fig:datapipeline}
\end{figure*}

The primary raw input data in our work are obtained from line-of-sight (LOS) magnetograms of ARs provided by the Helioseismic and Magnetic Imager (HMI) \cite{Schou2011} onboard the Solar Dynamics Observatory (SDO) \cite{Pesnell2011}, which are publicly available as a data product named Spaceweather HMI Active Region Patches (SHARP) \cite{Bobra2014} from the Joint Science Operations Center\footnote{http://jsoc.stanford.edu} at a temporal cadence of 12 minutes. In this work, we utilized magnetograms spanning from May 2010 to December 2018, sampling magnetograms at a cadence of one hour. The magnetograms of AR patches contain rasters of magnetic field strength values typically ranging from $\sim$$\pm$4500 G. An example of magnetogram of an AR patch is shown in Fig.~\ref{fig:dataexample} (a). Along with magnetograms, we use bitmaps (another data product from the SHARP series)  which define the region with pixels located within or outside the ARs, providing the region of interest within the AR patch as shown in Fig.~\ref{fig:dataexample} (b). The bitmaps are equal in size to the LOS magnetograms of AR patches, where the area represented by white pixels shows the region within the AR and hence our region of interest \cite{Bobra2014}. For each AR patch, we assign a binary label using peak X-ray flux converted to NOAA/GOES flare classes such that: (i) $\geq$M indicates Flare (FL) signifying the existence of a relatively strong flaring activity, and (ii) $<$M indicates No Flare (NF) with a prediction window of 24 hours. To illustrate, from the timestamp of an AR patch to the next 24 hours, if the maximum NOAA/GOES flare class is $<$M, then we label the AR patch as NF; otherwise, FL.

\begin{figure*}[htbp!]
\centering
\includegraphics[width=0.98\linewidth ]{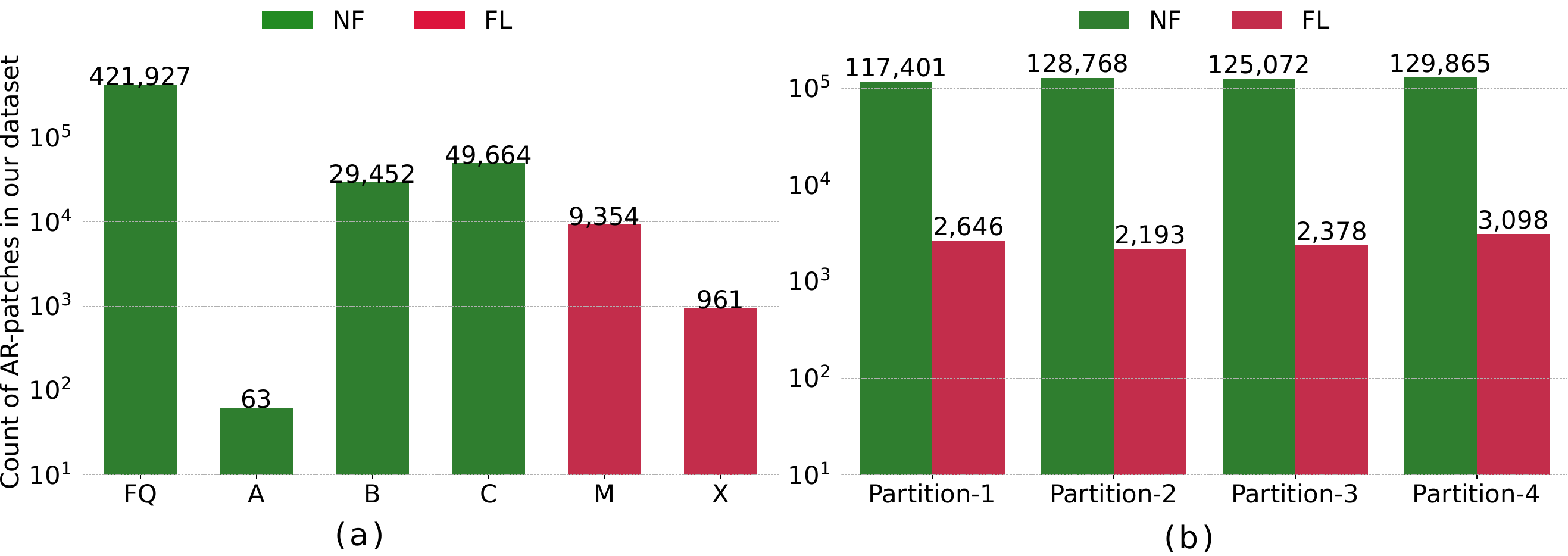}
\caption[]{(a) A bar plot representing the overall distribution of flare classes in our dataset (b) A bar plot showing binarized ($\geq$M) flare distributions across the four temporally non-overlapping tri-monthly data partitions. Note: The height of the bars are in logarithmic scale.}
\label{fig:data}
\vspace{-10pt}
\end{figure*}

In our data processing pipeline, introduced in our prior work \cite{pandeyecml2024} and illustrated in Fig.~\ref{fig:datapipeline}, we begin by collecting hourly instances of raw input magnetograms of AR patches, alongside their corresponding bitmaps. Our initial step involves applying the bitmap as a filter to precisely crop the AR patches, isolating the regions with high activity. Subsequently, we implement a size filter: if the resulting cropped AR patches are smaller than 70 pixels in width, we exclude them from our dataset. It is worth noting that we determine this threshold based on the overall data distribution, ensuring retention of all instances corresponding to 'FL' instances while removing those from the 'NF' class. Following this filtering stage, we proceed to adjust the magnetic flux. We cap the flux values at $\pm$256G, and any flux values within $\pm$25G are set to 0 to mitigate noise. Ensuring uniformity in size, we apply zero-padding to patches smaller than 512$\times$512 pixels. Conversely, for larger patches, exceeding 512$\times$512 pixels, we employ a 512$\times$512 kernel to select the patch with the maximum total unsigned flux (namely USFLUX, which is the sum of the absolute value of magnetic field strength represented as raster values in magnetograms). By doing this, we aim to minimize information loss by picking a spatial window where the total flux is the highest, which is more likely to include the regions of interest. Finally, to standardize the representation, all patches are scaled to fit within the range of 0-255, facilitating the generation of images. An example of a final processed image is shown in Fig.~\ref{fig:dataexample} (c) generated using the magnetogram of AR patch in Fig.~\ref{fig:dataexample} (a) and the corresponding bitmap in Fig.~\ref{fig:dataexample} (b), providing an illustration of the outcomes of our data preprocessing steps. 

The overall distribution of our labeled AR patches data, with binary flare classes NF (comprising flare-quiet (FQ), A-, B-, and C-class flares) and FL (including M- and X-class flares), is shown in Fig.~\ref{fig:data} (a). In total, we have 501,106 instances belonging to the NF class and 10,315 instances belonging to the FL class, resulting in a class imbalance ratio of $\sim$ 1:49. We split our dataset into four non-overlapping tri-monthly partitions as shown in Fig.~\ref{fig:data} (b), using the onset timestamps of the HARP series to ensure that each AR trajectory remains entirely within a single partition, thus avoiding any overlap.  This approach contrasts with the method described in \cite{pandey2021bigdata}, which uses magnetogram observation timestamps for partitioning the full-disk magnetograms. Finally, we use Partitions 1 and 2 as our training set while Partitions 3 and 4 are used as validation and test set respectively. The preprocessed dataset used in this study is publicly available from \cite{data}.

\begin{figure*}[htbp!]
\centering
\includegraphics[width=\linewidth ]{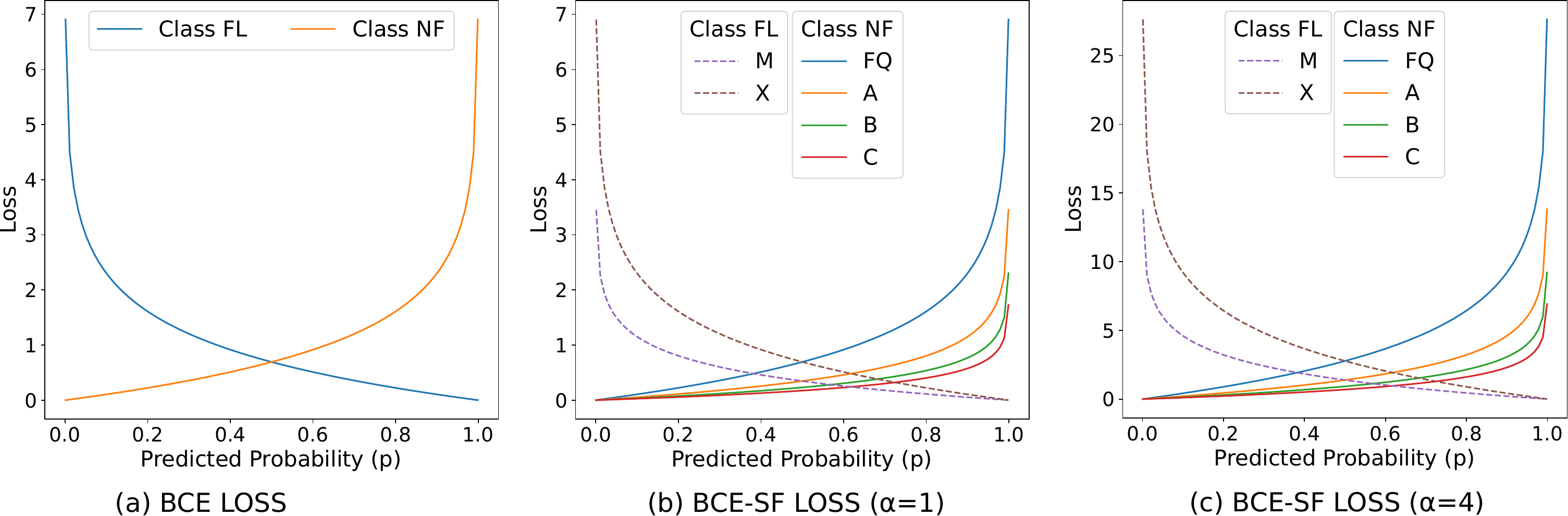}
\caption[]{An illustrative plot showing: (a) Standard binary cross-entropy (BCE) loss. (b-c) BCE for solar flare prediction (BCE-SF) which encodes ordinal flare characteristics as loss weighting mechanism with $\alpha$=1 and $\alpha$=4 respectively. Note: FL class indicates target 1 and NF class indicates target 0.}
\label{fig:loss}
\end{figure*}

The task of solar flare prediction in this work is formalized as a binary image classification problem; therefore, we select a general CNN based model, ResNet34 \cite{resnet}. Recently, attention-based models, notably Vision Transformers (ViTs) \cite{vit}, have gained prominence for their superior performance in image classification tasks. Despite their state-of-the-art results, these models tend to have a high number of trainable parameters, making them resource-heavy and less suitable for applications with limited computational resources or small datasets. For our specific application involving a small dataset, we opted for a more straightforward approach using a CNN-based model, specifically ResNet34. We modified the ResNet34 architecture to handle 1-channel input magnetogram images (grayscale images) by adding an initial convolutional layer with a 3$\times$3 kernel, a stride of 1, and three output feature maps. This modification allows the model to utilize pre-trained weights effectively while processing the 1-channel magnetogram images. The final architecture includes 34 convolutional layers (including all residual and basic convolutional layers), one max pooling layer, one adaptive average pooling layer, and one fully connected layer.

\section{Ordinality-aware Loss Function Design} \label{sec:method} 
In this work, we utilize a novel loss function designed for binary solar flare prediction that encodes the ordinal flare characteristics in the standard binary cross-entropy (BCE) loss. Let \( N \) be the total number of instances in a batch. Let \( y_i \) denote the true label for the \( i \)-th sample, where \( y_i \in \{0, 1\} \). Let \( p_i \) be the predicted probability that the \( i \)-th sample belongs to the "FL" class (target 1), defined as \( p_i = \sigma(\hat{y}_i) \), where \( \hat{y}_i)\) is the model output (logit) and \( \sigma \) is the sigmoid function, then the standard binary cross-entropy \( \text{BCE} (y, \hat{y}) \)  loss function is represented as shown in Eq.~(\ref{eq:bce}) and the corresponding loss function plot is shown in Fig.~\ref{fig:loss} (a).

\begin{equation}\label{eq:bce}
    \text{BCE}(y, \hat{y}) = -\frac{1}{N} \sum_{i=1}^N \left[ y_i \log(p_i) + (1 - y_i) \log(1 - p_i) \right]
\end{equation}

As mentioned earlier, in the binary setting of solar flare prediction where our chosen threshold is $\geq$M, the two binary classes are: (i) the NF-class including FQ-, A-, B-, and C-class instances, and (ii) the FL-class including M- and X- class instances. As each of these individual flare classes are ordinal in nature, where FQ$<$A$<$B$<$C$<$M$<$X, we introduce a weighting factor based on the corresponding sub-classes (flare classes within each binary class) such that instances belonging to these sub-classes are represented by weights $\beta$ as shown in Eq.~(\ref{eq:beta}). 
\begin{equation} \label{eq:beta}
\beta = \begin{cases}
10 & \text{if } \text{FQ} \\
10^2 & \text{if } \text{A} \\
10^3 & \text{if } \text{B} \\
10^4 & \text{if } \text{C} \\
10^2 & \text{if } \text{M}\\
10 & \text{if } \text{X}\\
\end{cases}
\end{equation}

The BCE-SF loss is designed in such a way that the incorrect predictions in the binary flare classes result in different losses based on the sub-class of the instances. To elaborate further, since an X-class flare is ten times more powerful than an M-class flare, even though both belong to class FL, the loss value for an X-class flare should be higher than that for an M-class flare. Similarly, a B-class flare in the NF class is ten times weaker than C-class flares in the same binary class. Therefore, incorrect predictions of these two classes should have different loss values. Thus, we used these ordinal weights ($\beta_i$) representing individual flare classes, and our proposed binary cross-entropy loss for solar flare prediction (BCE-SF) can be represented as shown in Eq.~(\ref{eq:bcesf}).

\begin{equation}\label{eq:bcesf}
    \text{BCE-SF}(y, \hat{y}) = -\frac{1}{N} \sum_{i=1}^N \alpha \times  \text{BCE}(y_i,  \hat{y}_i) \times \frac{1}{\log_{10}(\beta_i)}
\end{equation}

Here, $\alpha$ is a scaling factor that aligns the loss values with the scale of the corresponding BCE loss. Specifically, when $\alpha = 1$, the maximum loss value for an incorrectly predicted instance matches the scale of the BCE loss, as shown in Fig.~\ref{fig:loss} (b). In this case, the loss value scale for FQ- and X-class (the two extremes of the binary categories) aligns with the corresponding BCE loss, while all other incorrect predictions have lower loss values. Similarly, when $\alpha = 4$, the minimum loss value for an incorrectly predicted instance matches the scale of the BCE loss, as shown in Fig.~\ref{fig:loss} (c). Here, the loss value scale for C-class aligns with the BCE loss scale, while all other incorrect predictions have higher loss values. Therefore, we recommend the range of $\alpha \in \left[1, 4\right]$ which can be regarded as a hyperparameter for optimal performance. It is important to note that the BCE-SF loss, by leveraging intrinsic flare characteristics, offers a simple modification to the BCE loss without introducing new model-dependent parameters.

\begin{figure*}[htbp!]
\centering
\includegraphics[width=0.92\linewidth ]{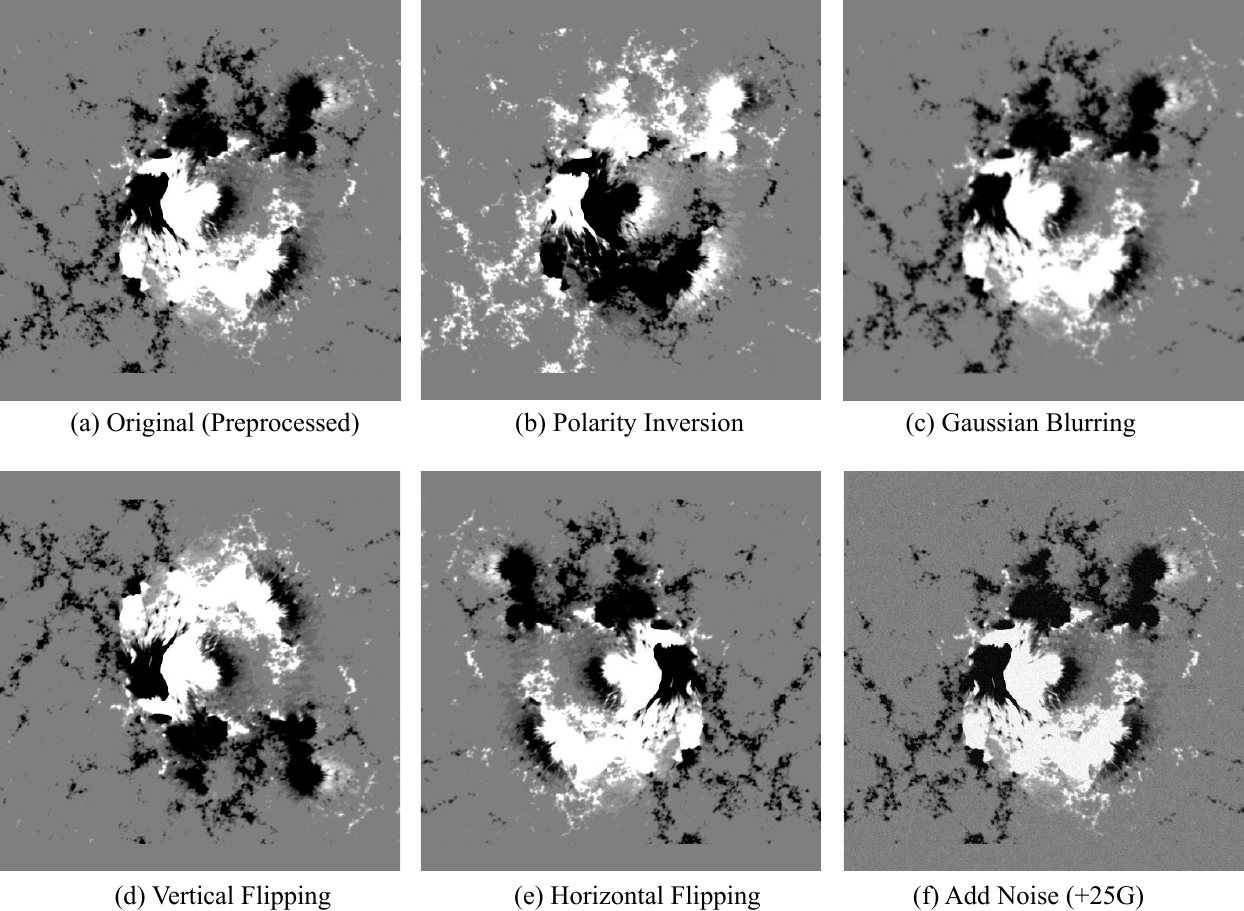}
\caption[]{An illustrative example of (a) input magnetogram of HMI AR patch corresponding to HARP number: 7115 (NOAA AR number: 12673). (b-f) five different augmentations applied to AR patch in (a). These augmentations are applied to the processed magnetograms before scaling to 0-255.}
\label{fig:aug}
\vspace{-10pt}
\end{figure*}

\section{Experimental Settings}\label{sec:expt}
In this section, we comprehensively delve into our dataset preparation methods for model training and evaluation, alongside detailing our model configurations in regard to the usage of BCE and BCE-SF loss functions, and hyperparameters. Furthemore, we provide the definition of our evaluation metrics and the rationale behind the selection of these metrics. 

\subsection{Dataset}\label{sec:exptsetup}

\begin{figure*}[htbp]
\centering
\includegraphics[width=0.92\linewidth ]{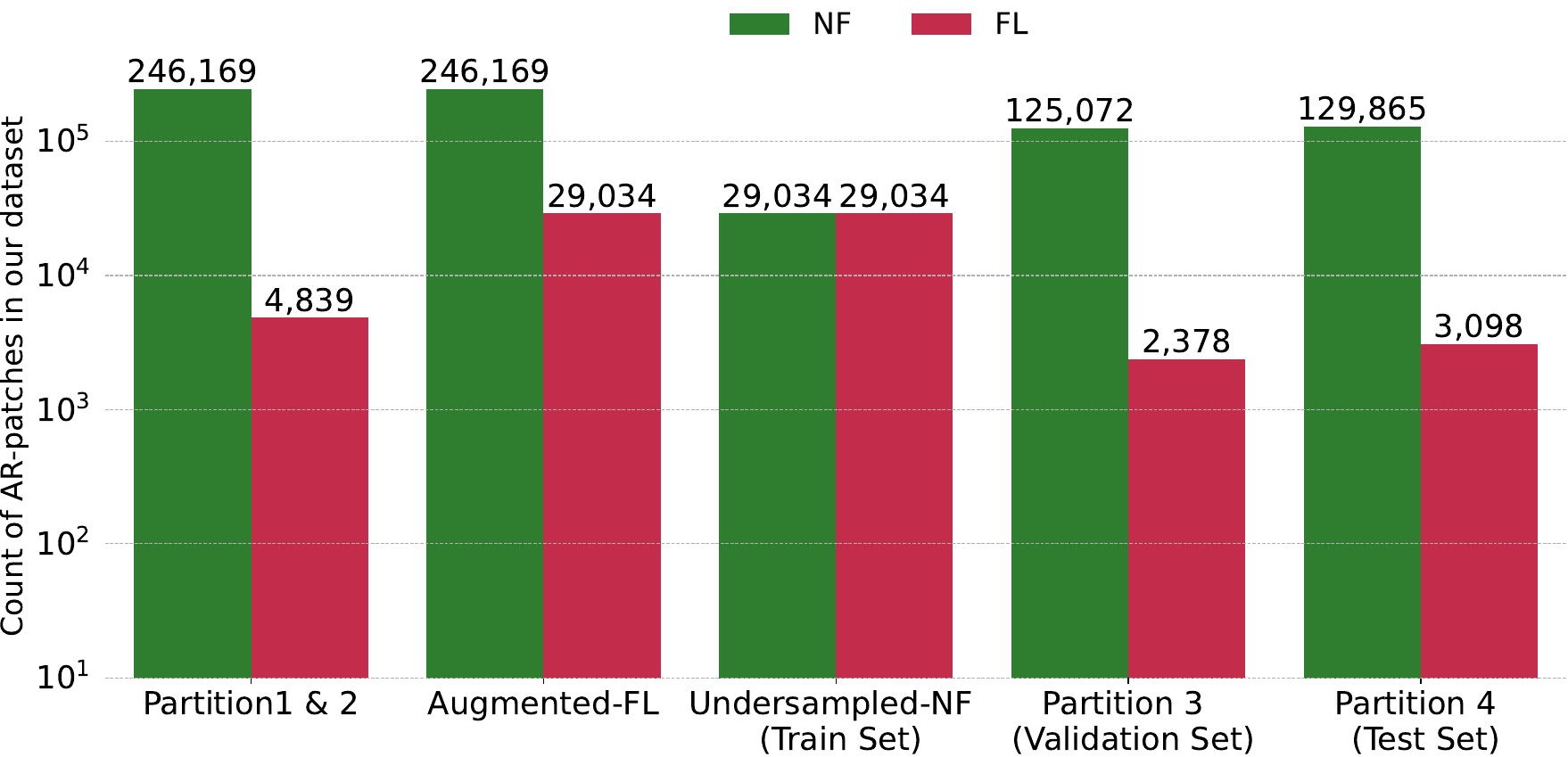}
\caption[]{The overall distribution of data instances partitioned into train set (showing original, augmented and undersampled data counts), validation set, and test set used in this work. }
\label{fig:dataset}
\vspace{-10pt}
\end{figure*}

As mentioned earlier in Sec.~\ref{sec:data}, we follow time-segmented trimonthly partitioning  to create four partitions of our entire dataset. Partition-1 and 2 combined are used as the training set. However, due to significant class imbalance in our dataset, we used undersampling together with data augmentation to create a balanced training set. Firstly, we augmented data instances belonging to the FL-class in our training set using five data augmentation techniques: (i) polarity inversion, which swaps the signs of positive polarity to negative and vice versa as shown in Fig.~\ref{fig:aug} (b), (ii) Gaussian filtering, which applies a Gaussian blur to the image to reduce noise and detail (Fig.~\ref{fig:aug} (c)), (iii) Vertical Flipping , which involves flipping the image along a horizontal axis (Fig.~\ref{fig:aug} (e)), (iv) Horizontal Flipping, which involves flipping the image along a vertical axis (Fig.~\ref{fig:aug} (d)), and (v) Adding random noise within $\pm$25G (Fig.~\ref{fig:aug} (f)). To balance the FL-Class instances with NF, we undersampled our training data by randomly selecting 30\% of instances belonging to A-, B-, C-class flares each, and $\sim$8\% of instances from FQ from both Partition-1 and 2. For realistic evaluation, we maintained the original imbalanced distribution in Partitions 3 and 4, which are our validation and test sets respectively, as shown in Fig.~\ref{fig:dataset}.

\subsection{Model Parameters}
\begin{table}[htbp]
\setlength{\tabcolsep}{6pt}
\renewcommand{\arraystretch}{1.5}

\caption{Hyperparameters search space with experimentally observed optimal hyperparameters for each model.}
\begin{center}
 \begin{tabular}{r r r r}
\hline
\multicolumn{2}{r}{}              
&                                            
\multicolumn{2}{c}{Optimal Parameters}\\
Hyperparameters & Search Space & BCE  & BCE-SF \\
\hline
Initial Learning Rate &  \{0.00001 to 0.01\}   &  0.01  & 0.001 \\

Weight Decay &  \{0.00001 to 0.01\}  &  0.01 & 0.001 \\

Batch Size & \{48, 64, 80\} & 64 & 64 \\ 

Scaling Factor ($\alpha$)& \{1, 2, 3, 4\} & N/A & 2 \\ 
\hline
\end{tabular}
\end{center}
\label{table:params}
\vspace{-5pt}
\end{table}
In our model hyperparameter selection process, we define the hyperparameter space, encompassing initial learning rates ($\eta$), weight decay parameters, batch sizes, and scaling factors ($\alpha$)  as shown in Table~\ref{table:params}. Following the definition of our hyperparameter space, we conduct a grid search across this space, evaluating on the validation set for all our models. During this search, we train our models using stochastic gradient descent (SGD) with BCE and BCE-SF loss function. Additionally, we employ a dynamic learning rate strategy called \texttt{ReduceLRonPlateau} with a factor of 0.3 and a patience period of 2 epochs. This learning rate scheduling mechanism starts the training with an initial learning rate ($\eta$) as mentioned in Table.~\ref{table:params}. If the validation loss does not improve for two consecutive epochs (patience period), the new learning rate is calculated as follows:
\[
\eta_{\text{new}} \text{ := } \eta_{\text{current}} \times \text{factor}
\]

Upon completing the grid search and evaluating the models, we identified the optimal hyperparameters as shown in Table~\ref{table:params}. These parameters exhibited superior performance during the search and we use these to train our final models for 50 epochs and evaluate on the test set.

\subsection{Evaluation Metrics}

True Skill Statistic (TSS, in Eq.~\ref{eq:TSS}) and Heidke Skill Score (HSS, in Eq.~\ref{eq:HSS}), derived from the four elements of confusion matrix: TP, TN, FP, FN are the two forecast skills scores widely used in evaluating flare prediction models. 

\begin{equation}\label{eq:TSS}
    TSS = \frac{TP}{TP+FN} - \frac{FP}{FP+TN} 
\end{equation}
\begin{equation}\label{eq:HSS}
    HSS = 2\times \frac{TP \times TN - FN \times FP}{((P \times (FN + TN) + (TP + FP) \times N))}
\end{equation}
\begin{center}
    \vspace{2pt}
where, $N = TN + FP$ and  $P = TP + FN$.    
\end{center}

The values of TSS and HSS range from -1 to 1, where 1 indicates all correct predictions, -1 represents all incorrect predictions (also, it means that all inverse predictions are correct, i.e., there is a skill), and 0 represents no skill. Unlike TSS, HSS is a metric that accounts for class imbalance. It is commonly used in evaluating solar flare prediction models because these datasets typically have a high imbalance ratio as discussed in \cite{Ahmadzadeh2019, Ahmadzadeh2021}. However, choosing a candidate model based on two skill scores becomes difficult, as it demands preference of one metric over another at the end. Therefore, by combining TSS and HSS in a geometric mean as in the Composite Skill Score (CSS, in Eq.~\ref{eq:CSS}), we obtain a single metric that balances between discrimination ability and imbalance awareness. 

\begin{equation}\label{eq:CSS}
CSS = 
\begin{cases}
0, & \text{if } TSS \times HSS < 0 \\
\sqrt{TSS \times HSS}, & \text{otherwise}
\end{cases}
\end{equation}

CSS considers both the discrimination power of the model (TSS) and its ability to outperform random chance (HSS), offering a more comprehensive evaluation. It provides a single metric that accounts for both aspects of model performance, making it more suitable for assessing forecast models, particularly in scenarios with class imbalance. Hence, we evaluate and compare our models based on the single metric, which is CSS but report both TSS and HSS for completeness. For reproducibility, the source codes for this work is publicly available from our open source repository \cite{ordinalloss_sf}.

\section{Experimental Evaluation} \label{sec:result}
\begin{figure*}[htbp]
\centering
\includegraphics[width=0.98\linewidth ]{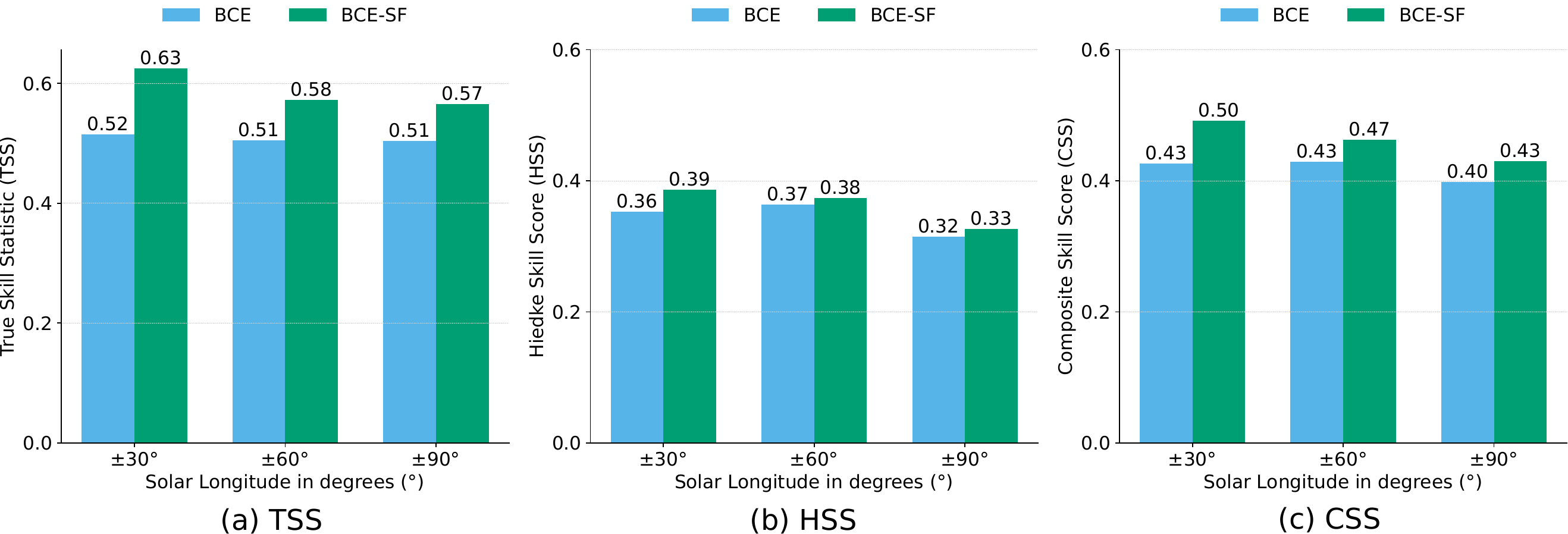}
\caption[]{Performance of our models trained with both BCE and BCE-SF loss was evaluated on the \textbf{test set} in terms of (a) TSS, (b) HSS, and (c) CSS. The skill scores are shown for different ranges of solar longitudes: $\pm$30$^\circ$, $\pm$60$^\circ$, and $\pm$90$^\circ$. This longitudinal ranges indicates the performance evaluated on ARs from the test set located within 0$^\circ$ to 30$^\circ$, 0$^\circ$ to 60$^\circ$, and 0$^\circ$ to 90$^\circ$ in both the directions (East (-ve) and  West (+ve)) of the Sun.
}
\label{fig:comparision}
\vspace{-10pt}
\end{figure*}

As explained earlier in Sec.~\ref{sec:exptsetup}, we conducted experiments to predict solar flares in a binary setting ($\geq$M-class flares) using a "train-validation-test split" of our entire dataset, which consists of magnetograms of AR patches covering a solar longitudinal range of $\pm$90$^\circ$ (i.e., the entire solar disk). We utilized the validation set to monitor the models' performance every epoch and tuned hyperparameters to optimize the CSS. After training the model with optimal hyperparameters, we employed a threshold tuning approach to calibrate our models by tuning the prediction score thresholds. This involved evaluating the performance of each model on the validation set at different threshold values ranging from 0.01 to 0.99 with an increment of 0.01. We selected the threshold that optimized CSS for each model individually. However, this resulted into a same threshold value of 0.69 for both the models trained with BCE and BCE-SF loss functions. These thresholds were then applied to the test set for both models, and the models' performance was reported on the test set. 

Additionally, we assessed the performance of both of our models on subsets of data representing different longitudinal coverages: within $\pm$30$^\circ$, $\pm$60$^\circ$, and $\pm$90$^\circ$ of solar longitude. To elaborate, longitudinal coverage of $\pm$30$^\circ$ indicates AR patches in our test set that correspond to the central region of the Sun, encompassing up to $-$30$^\circ$ in the East direction and up to $+$30$^\circ$ in the West direction from the center (0$^\circ$). A coverage of $\pm$60$^\circ$ extends further, encompassing a broader region that extends 60$^\circ$ to the East and West of the solar longitude. This range covers a significant portion of the Sun's surface, allowing for a more comprehensive examination of solar activity beyond just the central region but still within a reasonably close proximity to it. A coverage of $\pm$90$^\circ$ encompasses the entire test set.  The performance of our models relative to each other in terms of TSS, HSS, and CSS is illustrated in Fig.~\ref{fig:comparision} (a), (b), and (c), respectively.

Upon evaluation, we noted that the model trained with the BCE-SF loss function consistently outperformed the one optimized with BCE loss across all three skill scores and longitudinal coverages. Comparing the performance on the longitudinal coverage of $\pm$30$^\circ$, we observed $\sim$11\%, $\sim$3\%, and $\sim$7\% higher skill scores in terms of TSS, HSS, and CSS, respectively, as shown in Fig.~\ref{fig:comparision}. Furthermore, this improvement was also observed with increased longitudinal coverage. For ARs in the test set within $\pm$60$^\circ$, we noted $\sim$7\%, $\sim$1\%, and $\sim$4\% higher TSS, HSS, and CSS, respectively, with BCE-SF compared to the model trained with BCE. Overall, for our entire test set, we observed $\sim$6\%, $\sim$1\%, and $\sim$3\% higher scores in terms of TSS, HSS, and CSS, respectively, for the BCE-SF trained model over BCE, as illustrated in Fig.~\ref{fig:comparision}, indicated by $\pm$90$^\circ$. Additionally, our analysis revealed a linearly decreasing trend in model performance with increasing longitudinal coverage of ARs, with highest skill scores noted for ARs within ($\pm$30$^\circ$) and lowest when  within $\pm$90$^\circ$) for both of the models (BCE and BCE-SF).

\begin{figure*}[htbp]
\centering
\includegraphics[width=0.98\linewidth ]{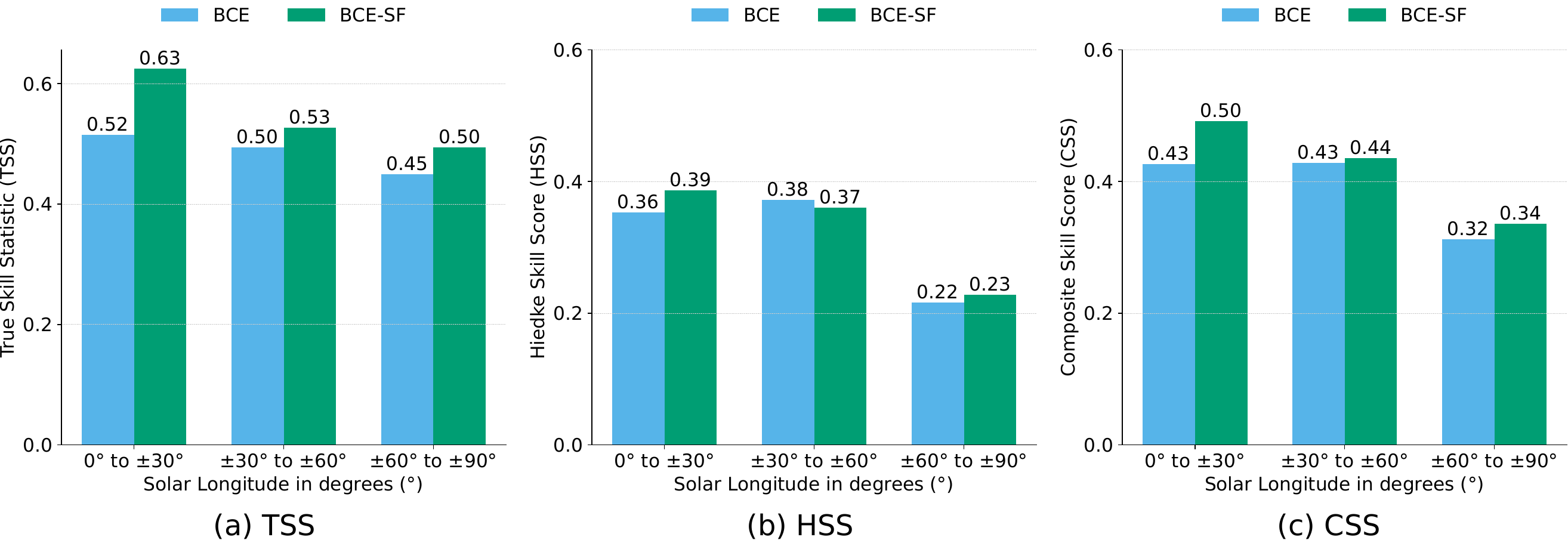}
\caption[]{Performance of our models trained with both BCE and BCE-SF loss was evaluated on the \textbf{test set} in terms of (a) TSS, (b) HSS, and (c) CSS. The skill scores are shown for different ranges of solar longitudes with \textbf{non-overlapping} regions: 0$^\circ$ to $\pm$30$^\circ$, $\pm$30$^\circ$ to $\pm$60$^\circ$, and $\pm$60$^\circ$ to $\pm$90$^\circ$. This longitudinal ranges indicates the performance evaluated on ARs from the test set located within 0$^\circ$ to 30$^\circ$, 30$^\circ$ to 60$^\circ$, and 60$^\circ$ to 90$^\circ$ in both the directions (East (-ve) and  West (+ve)) of the Sun.}
\label{fig:zone}
\vspace{-10pt}
\end{figure*}

After noticing the pattern indicating a decline in model performance with increasing longitudinal coverage, we investigated the effectiveness of our models on non-overlapping regions of solar longitudes. To facilitate this analysis, we delineated three zones: (i) within 0$^\circ$ to $\pm$30$^\circ$, (ii) the region between $\pm$30$^\circ$ to $\pm$60$^\circ$, and (iii) the region between $\pm$60$^\circ$ to $\pm$90$^\circ$. Similar to our earlier evaluation across overlapping longitudinal ranges, we computed all three skill scores to evaluate the model's performance across these zones. In doing so, we observed a similar linearly decreasing trend in skill scores as our earlier evaluation, highest in central regions (0$^\circ$ to $\pm$30$^\circ$) and lowest in limb regions ($\pm$60$^\circ$ to $\pm$90$^\circ$), as illustrated in Fig.~\ref{fig:zone} from all three models. Interestingly, we observed that while TSS is 3\% lower, HSS was 1\% higher with the BCE-trained model compared to BCE-SF when evaluating within $\pm$30$^\circ$ to $\pm$60$^\circ$ of the solar longitude, as shown in Fig.~\ref{fig:zone} (b). This highlights using a composite skill score, as choosing the model based solely on TSS or HSS scores might lead to a false sense of good performance. 

It is worth noting that while existing models are typically designed to predict solar flares up to $\pm$60$^\circ$ of the solar longitude, our model demonstrates capability in the near-limb regions, namely the region between $\pm$60$^\circ$ to $\pm$90$^\circ$. Despite having lower skill scores compared to those in the central region, this study reveals a new capability that demonstrates skill in the near-limb region, thereby advancing solar flare prediction. This advancement underscores the significance of our research in extending predictive models for solar flares beyond central regions, thereby improving our understanding and forecasting capabilities for solar phenomena.
 
\section{Conclusion and Future Work}\label{sec:conc}
In this study, we primarily introduced a novel ordinality-aware binary loss function to optimize data-driven predictive models and demonstrated its effectiveness in improving predictive capabilities compared to the standard binary loss function, specifically in the application of solar flare prediction. Furthermore, using our data preprocessing pipeline, we utilized AR patches encompassing the limb-to-limb range of the Sun (i.e., $\pm$90$^\circ$) to build flare prediction models capable of forecasting solar flares of magnitude $\geq$ M-class. Upon evaluating the capability of our limb-to-limb models, the results show that we can satisfactorily predict flaring activity despite severe projection effects, although there is room for improvement (the skill is limited when compared to central locations). Additionally, the results show that shape-based features in magnetograms are effective for predicting solar flares even when the ARs are close to the limbs.

While full-disk models are developed to complement AR-based models in near-limb regions, they lack the ability to localize AR-specific predictions. We define this work as an important step towards fully integrating ARs into solar flare prediction, with implications for advancing such predictions. Numerous avenues for future exploration exist, which include exploration on utilizing the actual peak X-ray fluxes as instance ordinality into the binary loss functions, investigating this approach with multi-modal solar observations, developing spatiotemporal models, and incorporating explanatory and interpretative frameworks into the model to enhance reliability.

\section*{Acknowledgments}
This work is supported in part under two NSF grants (Award \#2104004 and \#1931555) and a NASA SWR2O2R grant (Award \#80NSSC22K0272). The data used in this study is a courtesy of NASA/SDO and the AIA, EVE, and HMI science teams, and the NOAA National Geophysical Data Center (NGDC).

\bibliographystyle{IEEEtran}
\bibliography{new_bib}
\end{document}